\documentclass[10pt,twocolumn,letterpaper]{article}

 \usepackage{cvpr}              

\definecolor{cvprblue}{rgb}{0.21,0.49,0.74}
\usepackage[pagebackref,breaklinks,colorlinks,allcolors=cvprblue]{hyperref}

 \usepackage{booktabs}
 \usepackage{graphicx}
 \usepackage{multirow}
 \usepackage[export]{adjustbox}
\usepackage{soul}
\usepackage{amssymb}
\usepackage[dvipsnames]{xcolor}
\usepackage{caption}
\definecolor{delectricblue}{RGB}{93, 117, 131}
\colorlet{lightdelectricblue}{delectricblue!50} 
\definecolor{cambridgeblue}{rgb}{0.64, 0.76, 0.68}
\definecolor{bluegray}{rgb}{0.4, 0.6, 0.8}
\colorlet{delectblue}{bluegray!50}
\colorlet{delectgreen}{cambridgeblue!50}

\newcommand{\mysection}[1]{\vspace{2pt}}

\definecolor{mintgreen}{RGB}{193, 225, 193}
\definecolor{lightblue}{RGB}{173, 216, 230}
\definecolor{lightgreen}{RGB}{144, 238, 144}
\definecolor{lightyellow}{RGB}{255, 255, 204}
\definecolor{lightorange}{RGB}{255, 200, 150}
\definecolor{lightpurple}{RGB}{216, 191, 216}
\definecolor{lightcyan}{RGB}{224, 255, 255}
\definecolor{lightgray}{RGB}{230, 230, 230}

\usepackage{setspace}

\definecolor{cvprblue}{rgb}{0.21,0.49,0.74}
\usepackage[pagebackref,breaklinks,colorlinks,allcolors=cvprblue]{hyperref}

\def\paperID{19} 
\def\confName{CVPR}
\def\confYear{2025}

\begin{document}
\title{CLIP-SLA: Parameter-Efficient CLIP Adaptation for Continuous Sign Language Recognition}



\author{
\makebox[\textwidth][c]{%
    \parbox{17cm}{\centering
    Sarah Alyami$^{1,2}$ \quad Hamzah Luqman$^{1,3}$\\
    $^1$Information and Computer Science Department, King Fahd University of Petroleum and Minerals\\
    $^2$Computing Department, Imam Abdulrahman Bin Faisal University\\
    $^3$SDAIA–KFUPM Joint Research Center for Artificial Intelligence\\
    {\tt\small snalyami@iau.edu.sa \quad hluqman@kfupm.edu.sa}
    }
}
}


\twocolumn[{
\maketitle
\begin{center}
    \captionsetup{type=figure}  
    \begin{minipage}[b]{0.6\textwidth}  
        \centering
        \includegraphics[width=\linewidth]{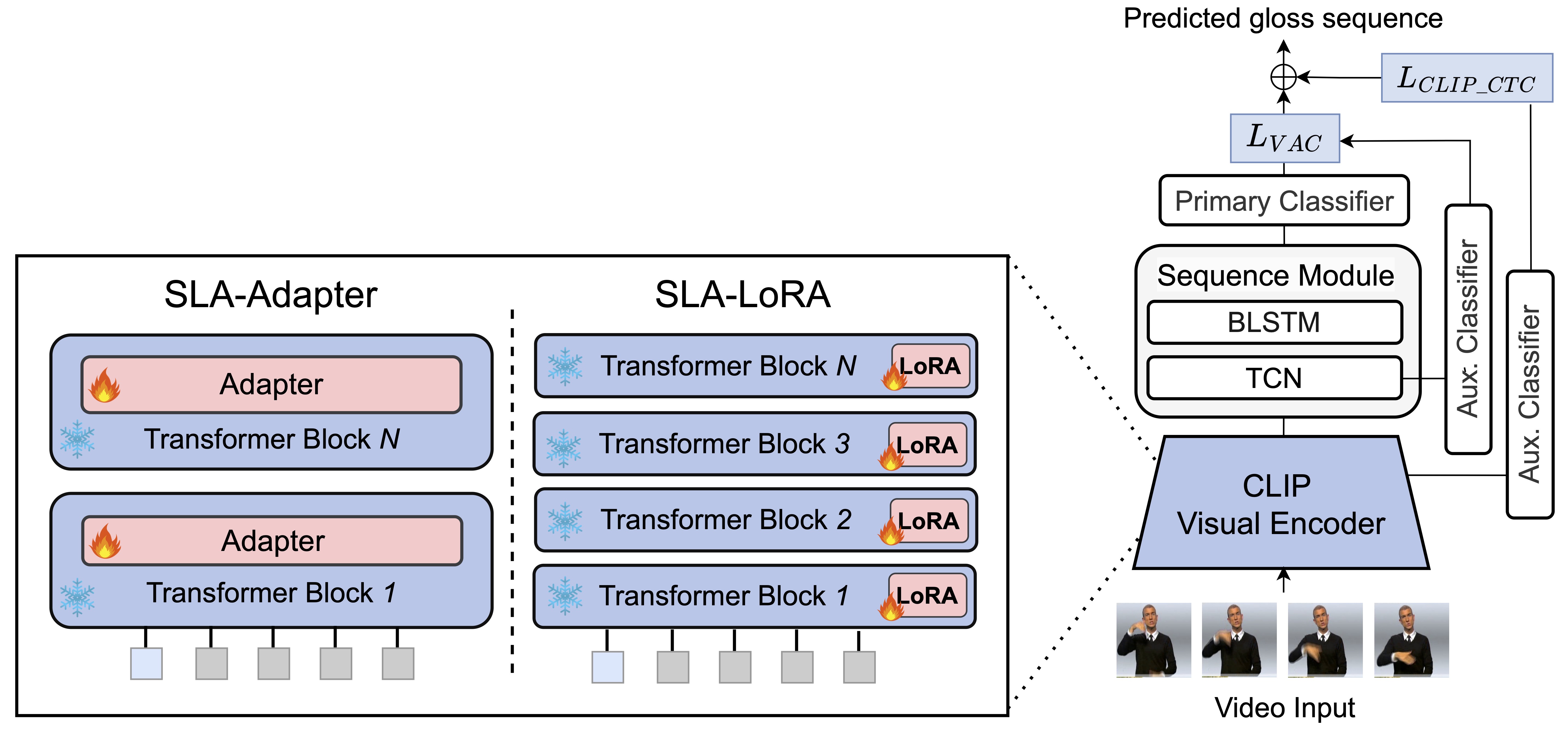}
        \caption*{(a)}  
    \end{minipage}
    \hfill
    \begin{minipage}[b]{0.38\textwidth}  
        \centering
        \includegraphics[width=\linewidth]{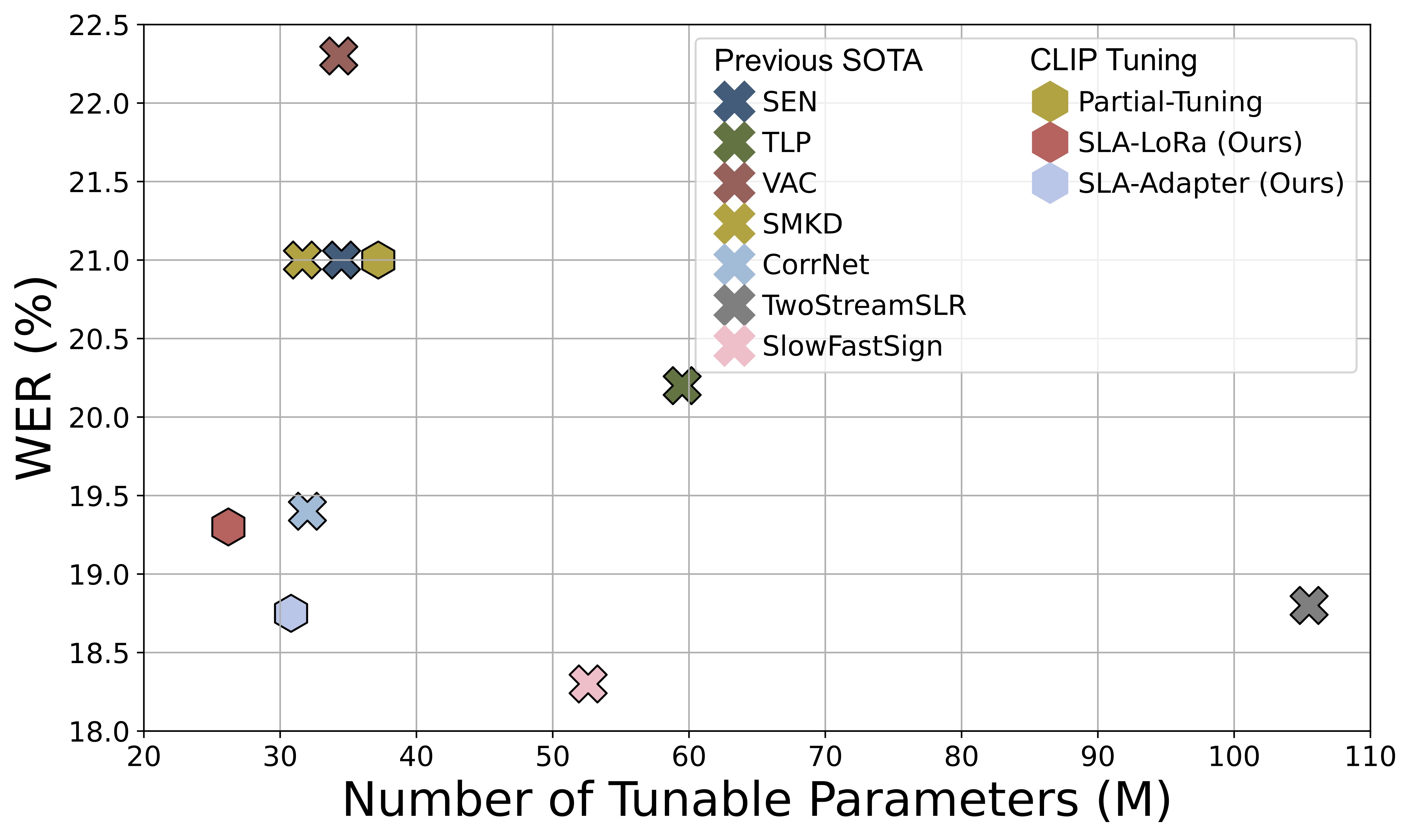}
        \caption*{(b)} 
    \end{minipage}
    \vspace{0.5em}  
    \caption{(a) The \textbf{CLIP-SLA} framework with two variants \textbf{SLA-Adapter} and \textbf{SLA-LoRA}. Both models leverage PEFT methods to transfer knowledge from the powerful CLIP pre-trained visual encoder to CSLR tasks efficiently. (b) A comparison between our CLIP-SLA model and state-of-the-art CSLR frameworks on the Phoenix2014 dataset, plotted against the number of tunable parameters.}
    \label{fig:abstract_figure}
\end{center}
}]




\begin{abstract}
Continuous sign language recognition (CSLR) focuses on interpreting and transcribing sequences of sign language gestures in videos. 
 In this work, we propose CLIP sign language adaptation (CLIP-SLA), a novel CSLR framework that leverages the powerful pre-trained visual encoder from the CLIP model to sign language tasks through parameter-efficient fine-tuning (PEFT). We introduce two variants, SLA-Adapter and SLA-LoRA, which integrate PEFT modules into the CLIP visual encoder, enabling fine-tuning with minimal trainable parameters. The effectiveness of the proposed frameworks is validated on four datasets: Phoenix2014, Phoenix2014-T, CSL-Daily, and Isharah-500, where both CLIP-SLA variants outperformed several SOTA models with fewer trainable parameters. Extensive ablation studies emphasize the effectiveness and flexibility of the proposed methods with different vision-language models for CSLR. These findings showcase the potential of adapting large-scale pre-trained models for scalable and efficient CSLR, which pave the way for future advancements in sign language understanding. Code is available at \url{https://github.com/snalyami/CLIP-SLA}.



\end{abstract}    
\section{Introduction}
\label{sec:intro}

Continuous sign language recognition (CSLR) is crucial for bridging the communication gap between deaf and hearing communities by automatically translating sign language videos into text \cite{Wadhawan2021}. CSLR models depend on learning spatio-temporal data in video streams to generate gloss-based transcriptions. This process requires efficient encoding of both spatial features (e.g., hand shapes, facial expressions) and temporal dependencies across frames \cite{ALYAMI2024103774}. However, CSLR faces challenges such as data scarcity, requiring expert gloss annotations, lack of clear boundaries between signs, and high computational demands due to long video sequences \cite{Tao2024}.

As a weakly annotated task, CSLR provides sequence-level gloss annotations without explicit temporal boundaries, making frame-to-gloss alignment a core challenge in this task \cite{Aloysius2020}. CSLR models typically rely on visual backbones to extract spatial features and temporal modules to model sign transitions. Fine-tuning ImageNet-pretrained backbones like ResNet \cite{hu2023corrnet, Min2021, Hao2021} or vision transformers (ViT) \cite{zhang2023c2st} for CSLR is common but computationally expensive and prone to overfitting on small sign language data. Instead, parameter-efficient fine-tuning (PEFT) methods provide a scalable alternative to full model fine-tuning by significantly reducing the costs of tuning large pre-trained models while maintaining competitive performance \cite{Xing2024}. Several PEFT techniques have been proposed, showing promising results in adapting pre-trained models, such as LoRA (Low-Rank Adaptation) \cite{hu2021lora}, adapters \cite{clip-adapter}, and prompt tuning \cite{zhou2022learning}.

Vision-language models (VLMs), such as CLIP, offer promising potential by integrating visual and linguistic modalities \cite{Xing2024}. Contrastive language-image pretraining (CLIP) is a multi-modal vision and language model proposed for image captioning \cite{radford2021learning}. The model aligns visual and textual representations via a contrastive learning objective. While CLIP excels in multi-modal generalization \cite{zhao2023clip, shen2021much, Abdelfattah_2023_ICCV, Wu_2023_CVPR}, adapting it for CSLR is non-trivial due to its lack of temporal modeling \cite{st-adapter} and the shortage of labeled sign language data.



Recently, CLIP has been adapted for various video understanding tasks that rely on sampling a small set of frames, such as action recognition \cite{luo2022clip4clip, wang2023actionclip, Liu_2023_CVPR, st-adapter} and isolated sign language recognition \cite{jiang-etal-2024-signclip}. However, CSLR requires a more fine-grained understanding of both local and global temporal dependencies across dense and longer video sequences \cite{HU2024109903}, making these frame-sampling-based frameworks less suitable for the task. This challenge underscores the need for lightweight yet effective adaptation strategies that can efficiently model the complex temporal structure of CSLR videos.

In this work, we propose CLIP-SLA (CLIP sign language adaptation), a framework that leverages CLIP’s capabilities for CSLR through two PEFT-based models: SLA-LoRA and SLA-Adapter (\cref{fig:abstract_figure} (a)). These approaches integrate temporal modeling within CLIP’s visual encoder to effectively capture spatio-temporal dependencies. Our framework achieves strong performance on several benchmark datasets including Phoenix2014, Phoenix2014-T, and CSL-Daily. Additionally, we evaluate our models on a new, more diverse dataset, \textit{Isharah-500} for continuous Saudi sign language.  The proposed methods outperform several state-of-the-art (SOTA) models with fewer trainable parameters (\cref{fig:abstract_figure} (b)). Comprehensive ablation studies further validate the efficiency and robustness of CLIP-SLA for sign language understanding.


\section{Related Work}
\label{sec:related_work}

\vspace{2mm}\noindent\textbf{Parameter-Efficient Transfer Learning. }
VLMs have significantly advanced multi-modal learning by enabling unified visual-text representations \cite{radford2021learning}. Models such as CLIP \cite{radford2021learning}, FLAVA \cite{singh2022flava}, and BLIP \cite{li2022blip} leverage large-scale pre-training on diverse datasets to align images and text in a shared embedding space. Among these, CLIP has emerged as a widely used model due to its robust generalization across tasks, aligning visual and textual features via contrastive learning \cite{Xing2024}. Given its potential, CLIP has been adapted to new domains through PEFT techniques, including prompt tuning \cite{khattak2023maple}, weight approximation \cite{hu2021lora}, and adapter-based methods \cite{zhang2021tip,clip-adapter,st-adapter}.

Prompt tuning optimizes model performance for downstream tasks by appending learnable prompts to the input before encoding. This strategy improves the model's adaptability without modifying the core architecture \cite{zhou2022learning, khattak2023maple,he2024biefficient}. Weight approximation methods, such as low-rank adaptation (LoRA) \cite{hu2021lora}, introduce trainable low-rank matrices into specific layers that allow efficient fine-tuning with minimal additional parameters. CLIP-LoRA \cite{zanella2024low} extends the application of LoRA from language models to CLIP vision and text encoders to enhance image classification performance. Adapter-based methods add lightweight trainable modules to a frozen backbone \cite{houlsby2019parameter}. This enables task-specific tuning while preserving the model’s pre-trained knowledge \cite{clip-adapter,st-adapter,zhang2021tip}. Given that CLIP processes each frame independently and lacks inherent temporal modeling, recent research has focused on adapting it for video understanding tasks \cite{st-adapter, Liu_2023_CVPR,luo2022clip4clip, wang2023actionclip}. Existing approaches either integrate temporal modules within CLIP’s transformer layers \cite{st-adapter, Liu_2023_CVPR} or apply temporal modeling after CLIP’s visual feature extraction \cite{luo2022clip4clip, wang2023actionclip}. 


\vspace{2mm}\noindent\textbf{CSLR Methods. }
CSLR has advanced rapidly with deep learning, leveraging visual backbones like 3D CNNs \cite{chen2023twostream,zuo2022c2slr}, 2D CNNs \cite{HU2024109903,Hu2023multilingual,jang2022signing}, and ViTs \cite{li2022multi} to extract spatial features, while sequential models such as 1D convolutions \cite{Jiao_2023_ICCV,jang2023self,hu2023adabrowse}, RNNs \cite{Hu2023multilingual,HU2024109903}, and transformers \cite{cui2023spatial,zheng2023cvt,zuo2024improving} capture temporal dependencies \cite{ALYAMI2024103774}. Efforts to enhance CSLR focus on optimizing training and improving spatio-temporal feature extraction. VAC \cite{Min2021} enforces temporal consistency through auxiliary losses, while SMKD \cite{Hao2021} applies knowledge distillation to refine visual-contextual interactions. To mitigate limited data, methods incorporate cross-lingual videos \cite{wei2023improving} or leverage self-supervised pre-training, such as SignBERT+ \cite{hu2023signbert+}. Researchers have introduced correlation maps \cite{hu2023corrnet}, attention mechanisms \cite{hu2023self}, and multi-stream architectures \cite{slowfast2024,jang2023self} to enhance CSLR accuracy. Multi-modal approaches, such as TwoStreamSLR \cite{chen2023twostream} and STMC \cite{Zhou2021}, integrate keypoint heatmaps and RGB data, while MSTN \cite{li2022multi} combines graph convolutions and transformers.


Vision-language alignment for sign language understanding has gained attraction recently \cite{jiang-etal-2024-signclip,zheng2023cvt,zhou2023gloss}. CVT-SLR \cite{zheng2023cvt} employs variational contrastive alignment to integrate visual and linguistic contexts, while GFSLT-VLP \cite{zhou2023gloss} applies CLIP-inspired pre-training for gloss-free sign language translation. However, these methods often require extensive pre-training \cite{zhou2023gloss,jiang-etal-2024-signclip} and their performance remains modest due to data constraints \cite{zheng2023cvt}.

Instead of training task-specific models, we efficiently adapt pre-trained VLMs like CLIP for CSLR. Training a CLIP-like model from scratch would demand large data and significant computation, whereas lightweight adaptation leverages CLIP’s large-scale image-text knowledge efficiently. While CLIP adaptation is widely studied in other domains \cite{auty2023learning,Zhang_2024_CVPR,chen2024clip,st-adapter,zanella2024low}, its potential for CSLR remains under-explored. Our work addresses this gap by investigating efficient and scalable adaptation strategies to extend CLIP’s image-text pre-training to continuous sign language videos.
\section{Method}
\label{method}
In this section, we first introduce our proposed CLIP-SLA framework with two variants: SLA-LoRA and SLA-Adapter. The general framework is shown in \cref{fig:abstract_figure} (a). The proposed model employs efficient and lightweight adaptation mechanisms tailored to CSLR. Both variants adapt CLIP's \cite{radford2021learning} powerful visual encoder while keeping the majority of its parameters frozen, enabling effective representation learning for CSLR. 

The CLIP-SLA architecture comprises a frozen CLIP visual encoder with a ViT-B/16 backbone, followed by a CSLR sequence modeling module that consists of temporal convolutional networks (TConv) and two bidirectional long short-term memory (BLSTM) layers. The adopted sequence module has been established to effectively capture sequential dependencies in continuous sign language videos in several CSLR frameworks \cite{hu2023corrnet,Hao2021,Min2021,hu2022temporal,hu2023self}. The final spatio-temporal features are passed to a fully connected classification layer that predicts the gloss sequence.

To train the model, we adopt a multi-loss setup. The primary loss is the CTC loss computed over the output of the main classifier after the BLSTM layers. To further improve alignment between visual features and gloss sequences, we incorporate the Visual Alignment Constraint (VAC) loss \cite{Min2021}, which encourages consistency between the predicted glosses and visual representations. Additionally, we introduce an auxiliary classifier directly after the CLIP visual encoder to provide early supervision. This auxiliary branch is also trained with a CTC loss, denoted as $\mathcal{L}_{CLIP\_CTC}$, ensuring that the visual backbone receives meaningful gradients even in the early training stages.

The total loss is computed as:
\[
\mathcal{L}_{total} = \mathcal{L}_{VAC} + \mathcal{L}_{CLIP\_CTC}
\]
During inference, only the main classifier is used to generate the predicted gloss sequence, while the auxiliary branches are removed. This training strategy improves both convergence and generalization by enforcing stronger supervision across the model's layers.

\subsection{SLA-LoRA}
\begin{figure}[]
\centering
\includegraphics[width=\linewidth]{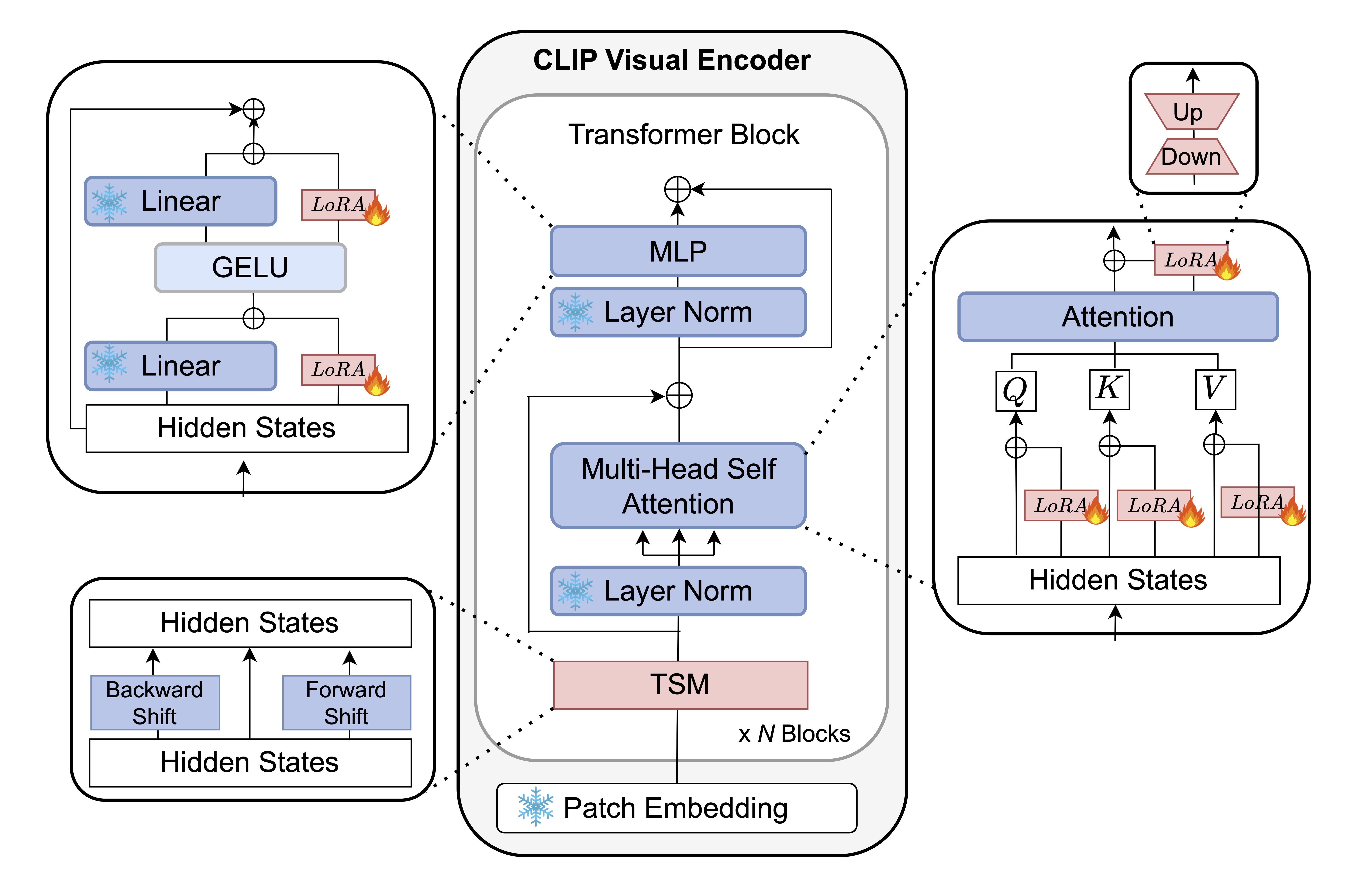}
\caption{The architecture of SLA-LoRA module. It shows the integration of the TSM and LoRA modules within the MHSA and MLP blocks of the ViT-based CLIP visual encoder.} 
\label{fig:sla-lora}
\end{figure}

SLA-LoRA is a lightweight CLIP adaptation framework that integrates the temporal shift module (TSM) \cite{lin2019tsm} with LoRA \cite{hu2021lora} to enhance the pre-trained CLIP visual encoder for CSLR. A detailed overview of the framework is shown in \cref{fig:sla-lora}. This framework consists of three key components: (1) TSM for temporal modeling, (2) LoRA applied to multi-head self-attention (MHSA) layers, and (3) LoRA applied to multi-layer perceptron (MLP) layers. By incorporating these elements, SLA-LoRA efficiently adapts the ViT-based CLIP encoder while maintaining the benefits of efficient tuning.

\vspace{2mm}\noindent\textbf{Temporal Shift Module (TSM). }
The CLIP encoder, pre-trained on large-scale image-text datasets, lacks temporal modeling, which is crucial for CSLR. While our framework includes temporal modules (TConv-BLSTM) after the visual backbone, we integrated TSM within each transformer layer to introduce temporal awareness early in the feature extraction process (see \cref{fig:sla-lora}). TSM \cite{lin2019tsm} has demonstrated strong performance in various video-related tasks \cite{wang2024temporal,an2023latent,tokenshift}. It is a lightweight and efficient approach that shifts a small portion of feature channels forward and backward along the temporal axis. This technique enables local temporal interactions without introducing additional parameters or significant computation overhead. Formally, given an input feature tensor \({X} \in \mathbb{R}^{B \times T \times L \times d}\), where \(B\) is the batch size, \(T\) is the temporal dimension, \(L\) is the number of spatial tokens, and \(d\) is the feature dimension, TSM works as follows:

\begin{equation}
    {X}'_{t, :, c} =
    \begin{cases}
        {X}_{t-1, :, c}, & 0 \leq c < \frac{d}{n_{\text{div}}} \\
        {X}_{t+1, :, c}, & \frac{d}{n_{\text{div}}} \leq c < \frac{2d}{n_{\text{div}}} \\
        {X}_{t, :, c}, & \text{otherwise}
    \end{cases}
    \label{eq:tsm}
\end{equation}

where ${X}'_{t, :, c}$ represents the updated feature tensor at frame $t$, 
covering all spatial tokens, and modifying the channel range $c$ based on the temporal shift.  The hyperparameter \(n_{\text{div}}\) controls how many channels participate in the temporal shift. In SLA-LoRA, we placed the TSM at the beginning of the transformer layer before the residual connection, as shown in \cref{fig:sla-lora}. This allows self-attention to operate on temporally-aware features while preserving the original residual pathway for stable learning. 

\vspace{2mm}\noindent\textbf{LoRA. }
LoRA \cite{hu2021lora} is an efficient tuning method that enables the adaptation of large pre-trained models with minimal additional parameters. SLA-LoRA leverages this technique by selectively injecting LoRA modules into the ViT architecture, specifically in the MHSA and MLP layers. Rather than fine-tuning all model parameters, LoRA introduces two learnable low-rank projection matrices, \( A \in \mathbb{R}^{r \times d} \) and \( B \in \mathbb{R}^{k \times r} \), to compute an update for the pre-trained weight matrix \( W \in \mathbb{R}^{d \times k} \). The modified transformation is computed as:

\begin{equation} 
h = W X + \Delta W X = W X + \frac{\alpha}{r} B A X
\label{eq:lora}
\end{equation}

where \( X \) is the input, \(\alpha\) is a scaling factor that controls the magnitude of the LoRA update, and \(r\) is the rank of the low-rank decomposition.

We utilized LoRA with the MHSA and MLP layers of our proposed framework, as shown in \cref{fig:sla-lora}. For the MHSA layers, LoRA modules are applied to the MHSA projections: query (\({W}_Q\)), key (\({W}_K\)), value (\({W}_V\)), and output projection (\({W}_O\)). These projections are key components of the attention mechanism, where the query, key, and value matrices determine attention weights, and the output projection aggregates results back into the original embedding dimension. By adapting MHSA, LoRA allows the model to capture sign language dependencies while retaining pre-trained knowledge, balancing efficiency and performance.

We also applied LoRA modules to the MLP layers to refine the extracted features and learn high-level interactions critical for CSLR tasks. These updates enable the model to capture complex transformations required for sign language recognition without disrupting the pre-trained weights. By integrating LoRA into both MHSA and MLP layers alongside TSM, SLA-LoRA effectively adapts the CLIP encoder to CSLR tasks with minimal overhead.

\subsection{SLA-Adapter}
Adapter-based fine-tuning is an effective method for adapting pre-trained models while preserving their generalization ability \cite{xing2024survey}. Adapters directly modify intermediate representations by introducing lightweight modules between layers, enabling stronger task-specific adaptation while maintaining the pre-trained model’s rich feature representations \cite{zhang2021tip,clip-adapter,st-adapter}. This approach is particularly beneficial for CSLR, where leveraging CLIP’s visual representations while integrating sign language-specific knowledge is essential for improved recognition performance. 

Typically, adapters consist of a down-projection linear layer, a non-linear activation function, and an up-projection linear layer. The feature matrix \({X} \in \mathbb{R}^{L \times d}\) is adapted as follows: 

\begin{equation}
\text{Adapter}({X}) = {X} + f({X} {W}_{\text{down}}) {W}_{\text{up}}, 
\label{eq:adapter}
\end{equation}

where \({W}_{\text{down}} \in \mathbb{R}^{d \times r}\) refers to the down-projection layer, \({W}_{\text{up}} \in \mathbb{R}^{r \times d}\) is the up-projection layer, and \(f(\cdot)\) is the activation function. A residual summation is applied to improve network learning stability. 

Our proposed SLA-Adapter fine-tunes the CLIP visual encoder for CSLR by selectively placing adapter modules within the ViT architecture to help in learning essential sign language features. Rather than fine-tuning the entire ViT model, only the adapter parameters are fine-tuned, ensuring efficient adaptation with minimal computational cost.

\vspace{2mm}\noindent\textbf{Adapter Design. }
Building on our approach in SLA-LoRA, we introduce temporal modeling early in the SLA-Adapter pipeline. 
However, instead of TSM, we integrate 3DConv adapters within CLIP's transformer layers. While 3DConvs are more computationally expensive than TSM, they offer a more effective way to capture local spatio-temporal dependencies and operate directly on feature representations without modifying the channel structure.

\begin{figure}[t]
    \centering
    \includegraphics[width=.8\linewidth]{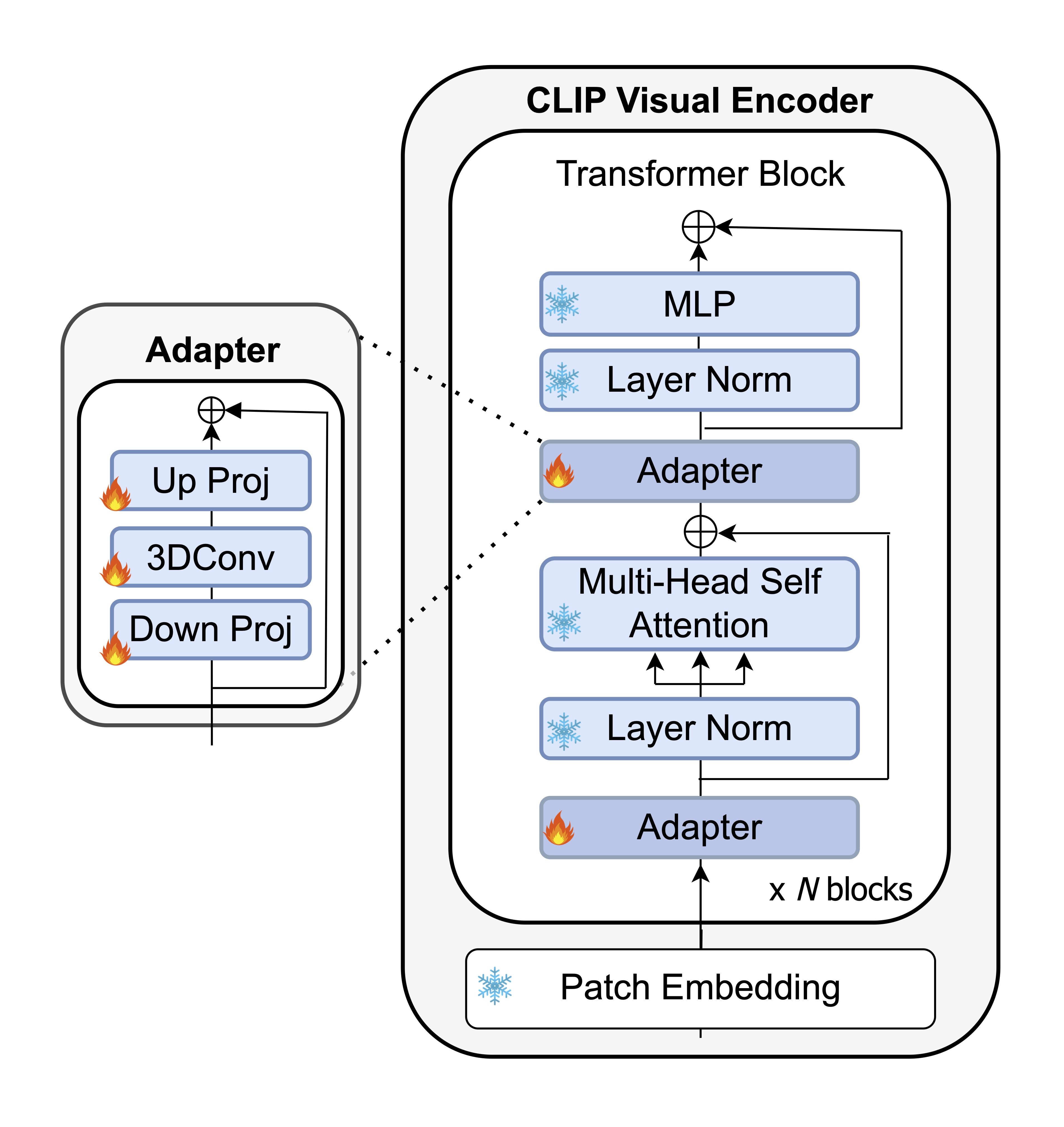}
    \caption{Overview of the proposed SLA-Adapter framework where the adapters are placed before the MHSA and MLP blocks. A detailed view of the time-aware adapter shows that the 3DConv layer is inserted between the downward and upward projections for effective spatio-temporal adaptation.
    }
    \label{fig:sla-adapter}
\end{figure}

As shown in \cref{fig:sla-adapter}, the time-aware adapter consists of down-projection layer, 3DConv, and up-projection layers. The down-projection layer reduces channel dimensions for computational efficiency, and the tokens are reshaped into a 3D structure for depth-wise 3DConv, which integrates temporal context with spatial features. The output is reshaped back to 2D and passed through an up-projection to restore the original dimensions, maintaining compatibility with the transformer. A residual connection adds the adapter’s output to the input tokens to preserve the pre-trained spatial representations while enhancing them with rich spatial-temporal correlations essential for CSLR. 

\vspace{2mm}\noindent\textbf{Adapter Placement. }
The placement of the adapter modules within the ViT backbone plays a crucial role in effectively adapting the pre-trained features to the target task \cite{st-adapter}. Similar to our approach with SLA-LoRA, we aim to adapt both MHSA and MLP components with the ViT backbone. Hence, we strategically place adapters in each transformer block before the MHSA layer and the MLP block, as shown in \cref{fig:sla-adapter}. This aims to ensure that task-specific adaptations are introduced at key stages of feature encoding. The adapters before the MHSA layer allow the model to inject task-specific dependencies early, enabling the self-attention mechanism to focus on relevant sign language spatial-temporal relationships. Similarly, placing adapters before the MLP block enhances the transformation of enriched features by refining them with task-specific nuances before further propagation. Similar to SLA-LoRA, we placed the first adapter before the residual connection, as shown in \cref{fig:sla-adapter}, to maintain the original residual pathway and ensure stable training.

\section{Experiments}
\begin{figure}[]
\centering
\includegraphics[width=\linewidth]{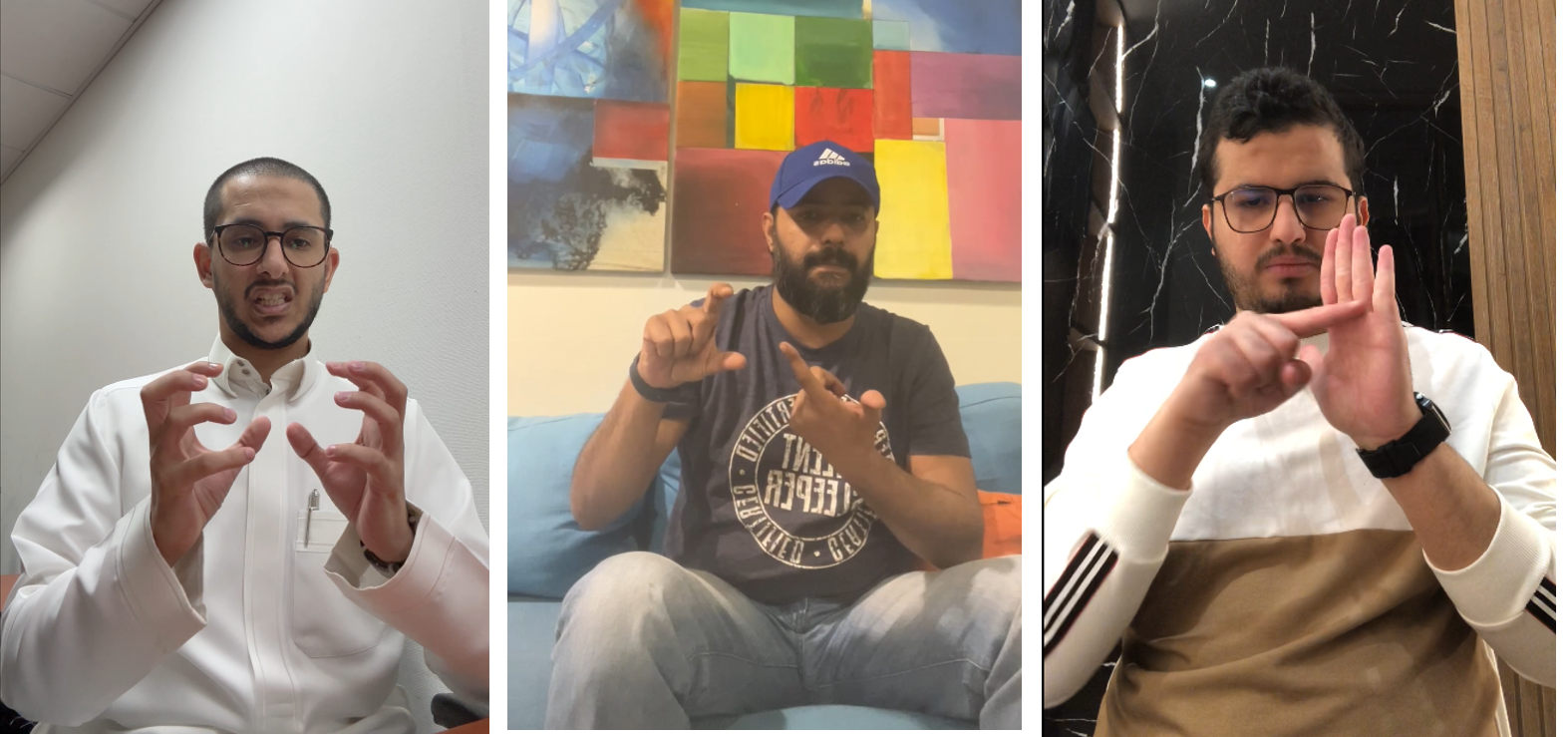}
\caption{Samples from the Isharah-500 dataset captured using smartphone cameras in unrestricted settings. } 
\label{fig:isharah_samples}
\end{figure}

\vspace{2mm}\noindent\textbf{Datasets. }
 The proposed framework has been evaluated on three standard benchmarking datasets Phoenix2014, Phoenix2014-T, and CSL-Daily. Moreover, we evaluated the robustness of the proposed framework on \textit{Isharah-500}, which is a new dataset collected in an ongoing project for Saudi sign language dataset development. Phoenix14 dataset \cite{Koller2015} includes recordings of German weather forecasts performed by 9 signers. It consists of 6,841 sentences representing 1,295 unique signs. Phoenix2014-T dataset \cite{Camgoz2018}, tailored for tasks in CSLR and sign language translation tasks, consists of 8,247 sentences spanning 1,085 signs. The CSL-Daily \cite{9578398} dataset focuses on daily life activities translated into Chinese sign language. The dataset consists of 20,654 videos, with a gloss vocabulary of 2,000. The Isharah-500 dataset is comprised of more challenging and realistic videos recorded using smartphone cameras in diverse conditions. These videos feature a variety of signers, backgrounds, lighting scenarios, and camera resolutions, as illustrated in \cref{fig:isharah_samples}. The dataset features 7,500 videos of sign language sentences with 388 unique signs performed by 15 fluent signers. It is divided into 5,000 videos for training samples, 500 videos for development, and 2,000 for testing. The dataset follows a signer-independent setup, with videos from 10 signers are used for the training set, while the development and test sets contain videos from the remaining 5 signers.

\vspace{2mm}\noindent\textbf{Training Details. }
The ViT-B/16 model with CLIP weights was used as the visual backbone. The framework was developed using PyTorch and the proposed model was optimized using Adam optimizer with \(10^{-4}\) weight decay and a batch size of two. SLA-LoRA models were trained for 35 epochs, while SLA-Adapter models were trained for 40 epochs.  An initial learning rate of \(10^{-4}\) was used which is reduced by a factor of 5 at the 20th and 30th epochs. During training, the frames were resized to 256x256 and then randomly cropped to 224x224. We also used random horizontal flipping and temporal rescaling techniques for data augmentation, while only center cropping was applied during inference. A beam decoder with 10 beams is utilized for decoding.  

\vspace{2mm}\noindent\textbf{Comparison with SOTA methods. }
We evaluate our approach using word error rate (WER), a widely adopted metric in CSLR research \cite{ALYAMI2024103774}. WER measures the discrepancy between predicted and ground truth sequences by calculating the minimum number of edits (insertions, deletions, and substitutions) required for alignment. To assess the effectiveness of our SLA-LoRA and SLA-Adapter models, we compare them against previous SOTA CSLR methods in \cref{tab:results_compare}. Additionally, we analyze the impact of our adaptation methods by benchmarking against other CLIP-based tuning methods, including zero-shot feature extraction (frozen CLIP visual encoder), partial fine-tuning (Partial FT) of the last two transformer blocks (11 and 12), and full fine-tuning (Full FT) of the entire CLIP visual encoder. As shown in \cref{tab:results_compare}, our PEFT-based models outperformed partial and naive full fine-tuning approaches. Full fine-tuning leads to weaker performance, likely due to catastrophic forgetting, where CLIP’s pre-trained knowledge is overwritten when all model weights are updated during fine-tuning. 

Compared to previous methods, both SLA-Adapter and SLA-LoRA achieve strong performance across Phoenix2014, Phoenix2014-T, and CSL-Daily, outperforming most RGB-based methods. On these datasets, SLA-Adapter achieves test WERs of 18.8\% and 19.5\%, and 25.8\% respectively, achieving comparable results to SlowFastSign \cite{slowfast2024}. SLA-LoRA obtained 19.3\%, 19.4\%, and 25.8\% test WERs on Phoenix2014, Phoenix2014-T, and CSL-Daily datasets, respectively. It also outperforms the majority of previous RGB-based methods. Notably, SLA-Adapter surpasses SLA-LoRA on Phoenix2014 with a 0.5 WER difference, which is likely due to it having more trainable parameters. However, SLA-LoRA achieves a lower WER on Phoenix2014-T (19.4\% vs. 19.5\%) and matches SLA-Adapter’s performance on CSL-Daily (25.8\%) while using 4.6M fewer tunable parameters, which highlights its efficiency in resource-constrained settings.


{As for the Isharah-500 dataset, both SLA-LoRA and SLA-Adapter demonstrated strong generalization on this challenging dataset, achieving test WERs of 24.0\% and 22.4\%, respectively. For comparison, we also evaluated the previous SOTA model, SlowFastSign \cite{slowfast2024}, on the same dataset, where SLA-LoRA and SLA-Adapter outperformed SlowFastSign significantly, achieving WER reductions of 33 and 35, respectively. These results validate the robustness of our frameworks in handling realistic and challenging scenarios, with CLIP's extensive knowledge and our specialized adaptations proving particularly effective in difficult cases, such as poor lighting and cluttered backgrounds encountered in the dataset's videos. }

\begin{table}[]
\centering
\caption{{Comparison with SOTA methods on Phoenix2014, Phoenix2014-T, and CSL-Daily. Bold and underlined indicate best and second-best results. The "Params (M)" column reports the total number of tunable parameters in each framework (in millions).}
   }
\label{tab:results_compare}
\footnotesize
\resizebox{\columnwidth}{!}{%

\begin{tabular}{lccccccc}
\toprule
                                            \multirow{2}{*}{\textbf{Method}}        &          \multirow{2}{*}{Params (M)}    & \multicolumn{2}{c}{Phoenix2014}                  & \multicolumn{2}{c}{Phoenix2014-T} & \multicolumn{2}{c}{CSL-Daily} \\
 & & Dev           & Test          & Dev             & Test            & Dev           & Test          \\ \midrule
\multicolumn{1}{l}{\textbf{Multi-Modal Methods}}       &             &               &               &                 &                 &               &               \\ \midrule 
C2SLR \cite{zuo2022c2slr}                            & NA          & 20.5          & 20.4          & 20.2            & 20.4            & 31.9          & 31.0          \\
CoSign \cite{Jiao_2023_ICCV}                         & 28.2        & 19.7          & 20.1          & 19.5            & 20.1            & 28.1          & 27.2          \\
SignBERTplus \cite{hu2023signbert+}                  & NA          & 19.9          & 20.0          & 18.8            & 19.9            & -             & -             \\
TwoStreamSLR \cite{chen2023twostream}                & 105.2       & 18.4          & 18.8          & 17.7            & 19.3            & 25.4          & 25.3          \\ \midrule
\multicolumn{1}{l}{\textbf{RGB-based Methods}}       &             &               &               &                 &                 &               &               \\ \midrule 
VAC \cite{Min2021}                                   & 34.3        & 21.2          & 22.3          & -               & -               & -             & -             \\
SMKD \cite{Hao2021}                                  & 31.6        & 20.8          & 21.0          & 20.8            & 22.4            & -             & -             \\
TLP \cite{hu2022temporal}                            & 59.5        & 19.7          & 20.8          & 19.4            & 21.2            & -             & -             \\
SEN \cite{hu2023self}                                & 34.5        & 19.5          & 21.0          & 19.3            & 20.7            & -             & -             \\
AdaBrowse \cite{hu2023adabrowse}                     & NA          & 19.6          & 20.7          & 19.5            & 20.6            & 31.2          & 30.7          \\
SSSLR \cite{jang2023self}                            & NA          & 20.9          & 20.7          & 20.5            & 22.3            & -             & -             \\
CTCA \cite{guo2023distilling}                        & NA          & 19.5          & 20.3          & 19.3            & 20.3            & 31.3          & 29.4          \\
CVT-SLR \cite{zheng2023cvt}                          &  NA          & 19.8          & 20.1          & 19.4            & 20.3            & -             & -             \\
CorrNet \cite{hu2023corrnet}                         & 32.0        & 18.8          & 19.4          & 18.9            & 20.5            & 30.6          & 30.1          \\ 
SlowFastSign \cite{slowfast2024}                     & 52.5        & \textbf{18.0}          & \textbf{18.3}          & \textbf{17.7}            & \textbf{19.3}            & \textbf{25.5}          & \textbf{24.9}          \\ \midrule
CLIP Frozen                                          & 23.1        & 21.1          & 28.6          & 28.6            & 26.9            & 27.7          & 35.3          \\
CLIP Partial FT                                      & 37.2        & 23.2          & 21.0          & 21.6            & 19.9            & 28.3          & 28.1          \\
CLIP Full FT                                         & 109.9       & 26.2          & 33.4          & 32.5            & 30.1            & 34.7          & 40.2          \\ \midrule
SLA-LoRa (ours)                                      & 26.2        & 19.7          & 19.3          & 19.8            & \textbf{19.4}   & \underline{26.0}          & \underline{25.8}          \\
SLA-Adapter (ours)                                   & 30.8        & \underline{18.5} & \underline{18.8} & \underline{18.8}   & 19.5            & {26.1} & \underline{25.8} \\ \bottomrule
\end{tabular}%

}
\end{table}

\vspace{2mm}\noindent\textbf{Efficiency Analysis.}  \cref{tab:efficiency_analysis} compares the training efficiency of our methods with the best-performing models, TwoStreamSLR \cite{chen2023twostream} and SlowFastSign \cite{slowfast2024}, on Phoenix2014. Our models achieve a balance between efficiency and accuracy, with SLA-Adapter matching TwoStreamSLR’s test WER and performing comparably to SlowFastSign (0.5 WER difference) while using significantly fewer parameters and training time. Nonetheless, given the relatively large visual backbone (ViT-B/16), our models remain computationally intensive despite being easier to train than fully fine-tuned CSLR models. Future work can explore model compression techniques like knowledge distillation or pruning to improve efficiency further.

\begin{table}[]
\centering
\caption{Efficiency analysis of our methods compared to previous SOTA CSLR frameworks.}
\footnotesize
\label{tab:efficiency_analysis}
\resizebox{\linewidth}{!}{%
\begin{tabular}{lccc}
\hline
\multirow{2}{*}{Method}               & \multirow{2}{*}{Params (M)} & \multirow{2}{*}{Training Epochs} & \multirow{2}{*}{Training Time (h)} \\
                                      &                             &                                  &                                    \\ \midrule
TwoStreamSLR \cite{chen2023twostream} & 105.2                       & 120                              & 240                                \\
SlowFastSign \cite{slowfast2024}      & 52.5                        & 80                               & 65.3                               \\ \midrule
SLA-LoRA (ours)                             & \textbf{26.2}               & \textbf{35}                      & \textbf{32.0}                      \\
SLA-Adapter (ours)                          & 30.8                        & 40                               & 36.6                               \\ \hline
\end{tabular}
}
\end{table}

\subsection{Ablation Studies}
Ablation studies are conducted on the three standard benchmark datasets to validate the effectiveness of the proposed CLIP-SLA framework and its performance under different configurations.  

\vspace{2mm}\noindent\textbf{TSM and LoRA Integration in SLA-LoRA.} 
In this section, we evaluate the effect of TSM and LoRA in the proposed framework and determine which layers of the backbone model should be adapted using LoRA. 
We first evaluated the contribution of TSM in the framework (\cref{fig:sla-lora}), removing TSM from the SLA-LorA framework results in a performance decline with an average increase of 0.7 WER across the three datasets. This highlights the role of TSM in efficiently capturing temporal dependencies by enabling information exchange across adjacent frames without introducing excessive computational overhead. Moreover, we observe that larger shift proportions $({n_{\text{div}}})$, such as 1/8 and 1/16, decreased the performance of the model while using 1/32 achieved a balance between spatial and temporal adaptation. 

We also examined the effect of applying LoRA only in the MHSA versus extending it to both MHSA and MLP blocks. Our results show that significant performance gains are achieved when LoRA is applied to both components, reducing WER by 1, 0.9, and 2 points on the Phoenix2014, Phoenix2014-T, and CSL-Daily datasets, respectively. This suggests that adapting the MLP block allows for better feature refinement for CSLR.

Moreover, we investigated which layers of the 12-layer ViT backbone should be adapted using LoRA. As shown in \cref{tab:lora_settings}, the best performance is achieved when LoRA is applied to layers 5-12, whereas extending LoRA to more layers results in a performance decline, likely due to disrupting the pretrained low-level features. Finally, we explored different LoRA ranks to determine the optimal rank settings. To minimize the hyper-parameter search, we set $\alpha$ (\cref{eq:lora}) with the same value of the rank in all experiments. As shown in \cref{tab:lora_settings}, our experiments show that higher ranks generally yield better performance, with the best results obtained at rank 16 for Phoenix2014 and rank 32 for Phoenix2014-T and CSL-Daily. 

\begin{table}[]
\centering
\caption{WERs (\%) of SLA-LoRA with different ranks and numbers of LoRA adapted layers within the 12-layered ViT backbone. }
\label{tab:lora_settings}
\resizebox{.9\columnwidth}{!}{%
\begin{tabular}{@{}lc|llllll@{}}
\toprule
                               \multicolumn{2}{l|}{LoRA Setting}       & \multicolumn{2}{l}{Phoenix14} & \multicolumn{2}{l}{Phoenix14-T} & \multicolumn{2}{l}{CSL-Daily} \\ 
                                  &    & Dev           & Test          & Dev            & Test           & Dev           & Test          \\ \midrule
\multirow{6}{*}{Layers} & 1-12 & 20.2          & 20.8          & 21.0             & 22.0             & 28.7          & 28.0            \\
                                  & 2-12 & 20.5          & 20.2          & 20.4           & 21.7           & 28.3          & 28.5          \\
                                  & 3-12 & 20.3          & 20.0            & 20.2           & 21.0             & 28.2          & 28.5          \\
                                  & 4-12  & 20.1          & 19.9          & 20.2           & 20.7           & 27.8          & 27.0            \\
                                  & 5-12  & \textbf{20.1}          & \textbf{19.8}          & \textbf{20.0}             & \textbf{20.6}           & \textbf{27.7}          & \textbf{26.8}          \\
                                  & 6-12  & 20.9          & 20.0            & 20.1           & 20.7           & 27.7          & 26.9          \\ \midrule
\multirow{4}{*}{ Rank}        & 4  & 20.1          & 19.8          & 20.0             & 20.6           & 27.7          & 26.8          \\
                                  & 8  & 19.8          & 19.6          & 19.9           & 20.1           & 27.5          & 26.5          \\
                                  & 16 & \textbf{19.7}          & \textbf{19.3}          & 19.8           & 20.0             & 27.3          & 26.3          \\
                                  & 32 & 19.7          & 20.0            & \textbf{19.8}           & {19.4}           & \textbf{26.0}            & \textbf{25.8}          \\ \bottomrule
\end{tabular}%
}
\end{table}

\vspace{2mm}\noindent\textbf{Adapter Settings in SLA-Adapter.}
Various adapter configurations were explored to evaluate their impact on the SLA-Adapter framework. We investigated applying a standard adapter \cite{clip-adapter} instead of the 3DConv adapter which obtained slightly lower performance with an average 0.5 WER increase across the three datasets. This demonstrates the effectiveness of the applied spatio-temporal CLIP adaptation provided by the 3DConv-based adapter for CSLR.
Next, we experimented with inserting the adapter only before the MHSA, which resulted in an average 0.3 WER performance decline, hence validating the need of adapting both MHSA and MLP features. 

Finally, we investigated the effect of the adapter's width and number of layers on the performance of CLIP-SLA model. We initially used adapters across all 12 layers and experimented with different adapter widths. As shown in \cref{tab:adapter_settings}. an adapter width of 256 resulted in the best performance across the three datasets. We then investigated how reducing the number of adapted layers affects the model performance. According to the obtained results, adapting more layers (starting from the 12th layer closer to the output) improves the model's performance. The best results were obtained when adapting 11 layers with Phoenix2014, 10 with Phoenix2014-T, and 8 with CSL-Daily. 

\begin{table}[]
\centering
\caption{WERs (\%) of SLA-Adapter with various configurations of adapter widths and layers adapted within the 12-layered ViT backbone.}
\label{tab:adapter_settings}
\resizebox{.9\columnwidth}{!}{%
\begin{tabular}{lc|llllll}
\toprule
\multicolumn{2}{l|}{Adapter Setting} & \multicolumn{2}{l}{Phoenix14} & \multicolumn{2}{l}{Phoenix14-T} & \multicolumn{2}{l}{CSL-Daily} \\
                           &        & Dev           & Test          & Dev            & Test           & Dev           & Test          \\ \midrule
\multirow{3}{*}{Width}     & 192    & 19.0            & 19.6          & 19.8           & 20.1           & 28.0            & 28.3          \\
                           & 256    & 18.6          & 18.9           & \textbf{18.9}           & \textbf{19.9}           & \textbf{27.3}          & \textbf{27.8}           \\
                           & 384    & \textbf{18.5} & \textbf{18.8}  & 19.0             & 20.0             & 27.8          & 28.0         \\ \midrule
\multirow{5}{*}{Layers}     & 1-12     & 18.5          & 18.8          & 18.9           & 19.9           & 27.3          & 27.8          \\
                           & 2-12             & 18.9          & 19.1           & 27.9          & 28.2   & 27.2          & 26.9        \\ 
                           & 3-12     & \textbf{18.5}          & \textbf{18.7}          & \textbf{18.8}           & \textbf{19.5}           & 27.2          & 27.8            \\
                           & 4-12      & 18.8          & 19.4          & 19.0             & 20.4           & 26.6          & 27.0            \\
                           & 5-12      & 19.1          & 19.5          & 19.5           & 20.7           & \textbf{26.1}          & \textbf{25.8} \\\bottomrule         
\end{tabular}%
}
\end{table}
\vspace{2mm}\noindent\textbf{Effect of CLIP CTC Loss.}
{We conducted an ablation study to assess the auxiliary CLIP classifier's impact on model performance. Removing it increased WER by 0.4 in SLA-LoRA and 0.7 in SLA-Adapter, confirming its role in providing direct supervision for the adapted CLIP encoder to ensure effective tuning. Combined with VAC loss, this multi-stage optimization enhances visual-context alignment, improving CSLR accuracy and robustness.}

\vspace{2mm}\noindent\textbf{Generalization to other VLMs.}
To evaluate the generalization of our adaptation methods, we applied SLA-LoRA and SLA-Adapter to another VLM, FLAVA \cite{singh2022flava}, which uses a ViT-B/16 backbone pre-trained with diverse objectives. As shown in  \cref{tab:flava}, Our adapters outperformed the frozen and partially fine-tuned settings of the same model, which demonstrates their robustness across VLMs. Additionally, we observe that CLIP achieves better results than FLAVA, likely due to its larger and more diverse pre-training data \cite{addepalli2024leveraging}. Nonetheless, these findings highlight the flexibility and potential of our methods in adapting diverse VLMs for CSLR. 

\begin{table}[]
\centering
\caption{WER (\%) results with adapting FLAVA visual encoder in our framework instead of CLIP model.}
\label{tab:flava}
\footnotesize
\resizebox{\columnwidth}{!}{%
\begin{tabular}{@{}lllllll@{}}
\toprule
                        & \multicolumn{2}{c}{Phoenix2014} & \multicolumn{2}{c}{Phoenix2014-T} & \multicolumn{2}{c}{CSL-Daily} \\ 
Setting                 & Dev        & Test      & Dev        & Test        & Dev       & Test      \\ \midrule
Frozen                  & 29.1           & 29.5            & 28.6           & 30.2              & 31.2            & 37.5                \\
Fine-tuning last layers & 21.7           & 22.4             & 23.3                         & 22.4                         & 29.5                         & 31.5                          \\ \midrule
SLA-LoRA (ours)              &      20.1     &      21.0      &             20.9    &    21.2             &          29.2     &         29.8      \\ 
SLA-Adapter (ours)              & \textbf{19.9}           & \textbf{20.1 }          &       \textbf{20.4 }         &    \textbf{21.0 }            &        \textbf{29.0}       &    \textbf{29.8}           \\ \bottomrule
\end{tabular}%
}
\end{table}


\subsection{Qualitative Results}
Samples from the Phoenix2014 dataset are analyzed to gain a deeper understanding of the performance of the two proposed methods. 

\begin{figure}[t]
    \centering
    \includegraphics[width=.9\columnwidth]{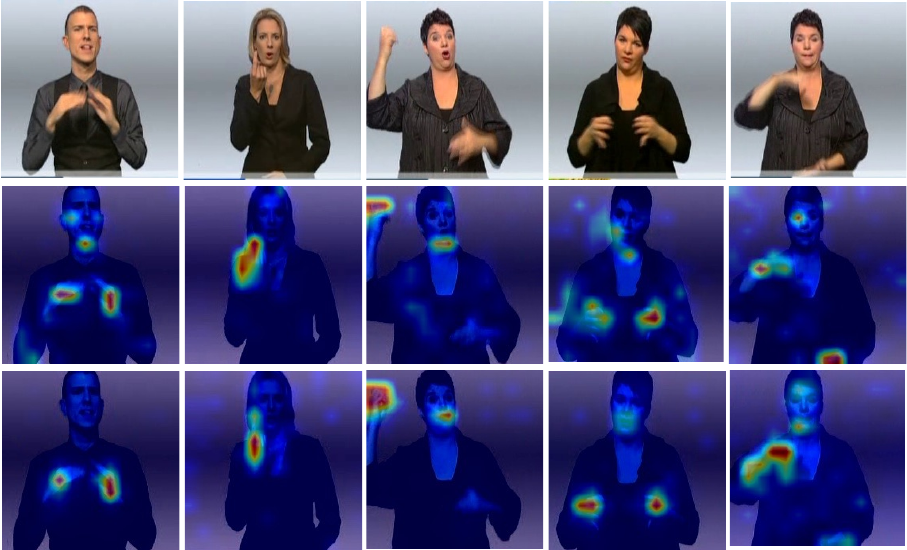}
    \caption{Visualizations of Grad-CAM from SLA-LoRA (2nd row) and SLA-Adapter (bottom row) showing focused attention to informative regions in sign language like hands and face. }
    \label{fig:gradcam_clip_sla}
\end{figure}

\vspace{2mm}\noindent\textbf{Attention Heatmaps. } 
Grad-CAM \cite{selvaraju2017grad} heatmaps generated by SLA-LoRA and SLA-Adapter using different test samples are displayed in \cref{fig:gradcam_clip_sla}. The generated heatmaps demonstrate that both approaches attend to critical regions for sign language understanding. The visualizations show that both models focus on the hands and mouth to capture hand shapes and mouthing cues, which are essential for interpreting signs. 

\vspace{2mm}\noindent\textbf{Gloss Predictions. }
Gloss predictions from SLA-LoRA and SLA-Adapter are shown in \cref{tab:gloss_preds_clip_sla}. In the first example, both models correctly recognized the sentence, demonstrating their ability to adapt CLIP for CSLR. The second example highlights a mistake by SLA-LoRA, which predicted "HAGEL" (hail) instead of "EINFLUSS" (influence), which can be attributed to the adaptation constraints in LoRA, where low-rank updates may not fully capture fine-grained sign-to-text mappings. The final example reveals errors in both models, where "FUENFZEHN" (fifteen) was misclassified as "VIERZEHN" (fourteen). This observation indicates that despite its advantages, our methods still face challenges in distinguishing subtle finger details required for accurate interpretation of fine-grained sign language details.

\begin{table}[t]
\centering
\caption{Gloss predictions of SLA-LoRA and SLA-Adapter. Errors are colored in pink.}
\label{tab:gloss_preds_clip_sla}
\resizebox{\linewidth}{!}{%
\begin{tabular}{ll}
\hline
\textbf{Ground Truth}       & \colorbox{lightblue}{TEMPERATUR} \colorbox{mintgreen}{NULL} \colorbox{lightyellow}{GRAD} \colorbox{lightorange}{KALT} \colorbox{lightpurple}{NORD} \colorbox{lightcyan}{MINUS} \colorbox{delectblue}{FUENF} \colorbox{lightgray}{GRAD}             \\
\textbf{SLA-LoRA} & \colorbox{lightblue}{TEMPERATUR} \colorbox{mintgreen}{NULL} \colorbox{lightyellow}{GRAD} \colorbox{lightorange}{KALT} \colorbox{lightpurple}{NORD} \colorbox{lightcyan}{MINUS} \colorbox{delectblue}{FUENF} \colorbox{lightgray}{GRAD}  \\ 
\textbf{SLA-Adapter} & \colorbox{lightblue}{TEMPERATUR} \colorbox{mintgreen}{NULL} \colorbox{lightyellow}{GRAD} \colorbox{lightorange}{KALT} \colorbox{lightpurple}{NORD} \colorbox{lightcyan}{MINUS} \colorbox{delectblue}{FUENF} \colorbox{lightgray}{GRAD}  \\ 
& \\  

\textbf{Ground Truth}       & \colorbox{lightblue}{MILD} \colorbox{mintgreen}{WEHEN} \colorbox{lightyellow}{ICH} \colorbox{lightorange}{RUSSLAND} \colorbox{lightpurple}{IX} \colorbox{lightcyan}{STARK} \colorbox{delectblue}{KOMMEN} \colorbox{lightgray}{EINFLUSS} \\
\textbf{SLA-LoRA}   & \colorbox{lightblue}{MILD} \colorbox{mintgreen}{WEHEN} \colorbox{lightyellow}{ICH} \colorbox{lightorange}{RUSSLAND} \colorbox{lightpurple}{IX} \colorbox{lightcyan}{STARK} \colorbox{delectblue}{KOMMEN} \colorbox{pink}{HAGEL} \\
\textbf{SLA-Adapter} & \colorbox{lightblue}{MILD} \colorbox{mintgreen}{WEHEN} \colorbox{lightyellow}{ICH} \colorbox{lightorange}{RUSSLAND} \colorbox{lightpurple}{IX} \colorbox{lightcyan}{STARK} \colorbox{delectblue}{KOMMEN} \colorbox{lightgray}{EINFLUSS}    \\

& \\  

\textbf{Ground Truth} & \colorbox{lightblue}{JETZT} \colorbox{mintgreen}{MORGEN} \colorbox{lightyellow}{WETTER} \colorbox{lightorange}{WIE-AUSSEHEN} \colorbox{lightpurple}{MORGEN} \colorbox{lightcyan}{FUENFZEHN} \colorbox{lightgray}{OKTOBER}  \\
\textbf{SLA-LoRA}   & \colorbox{lightblue}{JETZT} \colorbox{mintgreen}{MORGEN} \colorbox{lightyellow}{WETTER} \colorbox{lightorange}{WIE-AUSSEHEN} \colorbox{lightpurple}{MORGEN} \colorbox{pink}{VIERZEHN} \colorbox{lightgray}{OKTOBER} \\
\textbf{SLA-Adapter} & \colorbox{lightblue}{JETZT} \colorbox{mintgreen}{MORGEN} \colorbox{lightyellow}{WETTER} \colorbox{lightorange}{WIE-AUSSEHEN} \colorbox{lightpurple}{MORGEN} \colorbox{pink}{VIERZEHN} \colorbox{lightgray}{OKTOBER}  
\\  \bottomrule
\end{tabular}%
}
\end{table}

\section{Conclusion}
\label{sec:conclusion}

In this work, we introduced CLIP-SLA (CLIP Sign Language Adaptation), a framework that efficiently adapts CLIP for CSLR using PEFT techniques. Our approach addresses CLIP’s lack of temporal modeling and the scarcity of large-scale annotated sign language datasets. To tackle these challenges, we proposed SLA-LoRA and SLA-Adapter, which integrate temporal modeling into CLIP’s visual backbone while keeping computational costs low.

Experiments on Phoenix2014, Phoenix2014-T, CSL-Daily, and Isharah-500 show that both methods achieve strong performance with significantly fewer trainable parameters. Ablation studies further validate the effectiveness of integrating temporal modules and PEFT techniques for CSLR.

Beyond performance gains, our work highlights the potential of pretrained VLMs for CSLR, moving beyond conventional vision-based models. By leveraging lightweight adaptation methods, we show that vision-language knowledge can be transferred to CSLR without full fine-tuning. Future work could explore other PEFT techniques, such as prefix tuning and visual prompt tuning, or adapt emerging VLMs like LLaVA-OneVision \cite{li2025llavaonevision} and Molmo \cite{deitke2024molmo} to enhance sign language understanding through multi-modal learning.

\section*{Acknowledgments}
\noindent The authors would like to acknowledge the support received from the Saudi Data and AI Authority (SDAIA) and King Fahd University of Petroleum and Minerals (KFUPM) under the SDAIA-KFUPM Joint Research Center for Artificial Intelligence Grant {JRC-AI-RFP-14}.
{
    \small
    \bibliographystyle{ieeenat_fullname}
    \bibliography{main}
}


\end{document}